\def\year{2020}\relax
\documentclass[letterpaper]{article} % DO NOT CHANGE THIS
\usepackage{aaai20}  % DO NOT CHANGE THIS
\usepackage{times}  % DO NOT CHANGE THIS
\usepackage{helvet} % DO NOT CHANGE THIS
\usepackage{courier}  % DO NOT CHANGE THIS
\usepackage[hyphens]{url}  % DO NOT CHANGE THIS
\usepackage{graphicx} % DO NOT CHANGE THIS
\usepackage{graphicx}  % DO NOT CHANGE THIS
\usepackage{tabularx,xspace,multirow}
\usepackage{multicol,lipsum}
\usepackage{color}
\usepackage{xcolor}
\usepackage{colortbl,array} % für farbige cells
\usepackage{amsmath}
\definecolor{mygreen}{rgb}{0,0.6,0}
\definecolor{mygray}{rgb}{0.5,0.5,0.5}
\definecolor{mymauve}{rgb}{0.58,0,0.82}
\definecolor{myltgray}{rgb}{0.83,0.83,0.83}
\usepackage{courier}

\DeclareMathOperator*{\argmax}{arg\,max}

\newcolumntype{P}[1]{>{\centering\arraybackslash}p{#1}}
\newcolumntype{M}[1]{>{\centering\arraybackslash}m{#1}}

\title{Exploring Hyper-Parameter Optimization for Neural\\Machine Translation on GPU Architectures}
\author{Robert Lim\textsuperscript{\rm 1}\thanks{roblim1@cs.uoregon.edu}, Kenneth Heafield\textsuperscript{\rm 2}\thanks{kheafiel@inf.ed.ac.uk}, Hieu Hoang\textsuperscript{\rm 3}\thanks{hieuhoang@gmail.com}, Mark Briers\textsuperscript{\rm 3}\thanks{mbriers@turing.ac.uk}, Allen Malony\textsuperscript{\rm 1}\thanks{malony@cs.uoregon.edu}\\ % All authors must be in the same font size and format. Use \Large and \textbf to achieve this result when breaking a line
\textsuperscript{\rm 1} University of Oregon, Eugene, OR, U.S.A.\\
\textsuperscript{\rm 2} University of Edinburgh, Edinburgh, Scotland, U.K.\\
\textsuperscript{\rm 3} The Alan Turing Institute, London, England, U.K.\\
}

\begin{document}

\maketitle

\begin{abstract}
Neural machine translation (NMT) has been accelerated by deep learning neural networks over statistical-based approaches, due to the plethora and programmability of commodity heterogeneous computing architectures such as FPGAs and GPUs and the massive amount of training corpuses generated from news outlets, government agencies and social media.  Training a learning classifier for neural networks entails tuning hyper-parameters that would yield the best performance.  Unfortunately, the number of parameters for machine translation include discrete categories as well as continuous options, which makes for a combinatorial explosive problem. This research explores optimizing hyper-parameters when training deep learning neural networks for machine translation.  Specifically, our work investigates training a language model with Marian NMT.  Results compare NMT under various hyper-parameter settings across a variety of modern GPU architecture generations in single node and multi-node settings, revealing insights on which hyper-parameters matter most in terms of performance, such as words processed per second, convergence rates, and translation accuracy, and provides insights on how to best achieve high-performing NMT systems.
\end{abstract}

\section{Motivation}

The rapid adoption of neural network (NN) based approaches to machine translation (MT) has been attributed to the massive amounts of datasets, the affordability of high-performing commodity computers, and the accelerated progress in fields such as image recognition, computational systems biology and unmanned vehicles.  Research activity in NN-based machine translation has been taking place since the 1990s, but statistical machine translation (SMT) soared along with the successes of machine learning.  SMT incorporates a rule-based, data driven approach, and includes language models such as word based (n-grams), phrased-based, syntax-based and hierarchical based approaches.  Neural machine translation (NMT), on the other hand, does not require predefined rules, but learns lingusitic rules from statistical models, sequences and occurences from large corpuses.  Models trained using NNs produce even higher accuracy than existing SMT approaches, but training time can take anywhere from days to weeks to complete.  Suboptimal strategies are often difficult to find, given the dimensionality and its effect on parameter exploration.

One of the main difficulties of training neural networks is the millions of parameters that need to be estimated.  These parameters are estimated by optimization methods, such as stochastic gradient descent, where the solver seeks to identify the global optima.  Due to the combinatorial search space, local optimization in many cases is sufficent to generalize beyond the training set~\cite{Goodfellow-et-al-2016} (Ch. 8).  Thus, the tuning of hyper-parameters is paramount in accelerating training of neural networks.

In neural machine translation, modeling and training are crucial in achieving high performing systems.  A combination of hyper-parameter optimization methods to train a NMT system is investigated in this work.  Specifically, this work examines the stability of different optimization parameters in discovering local minima, and how a combination of hyper-parameters can lead to faster convergence.

The following contributions are made in this work:

\begin{itemize}
\item We identify which hyper-parameters matter most in contributing to the learning trajectory of NMT systems.
\item We analyze our findings for translation performance, training stability, convergence speed, and tuning cost.
\item We tie in systems execution performance with hyper-parameters.
\end{itemize}

\section{Related Work}

Hyper-parameter optimization has been an unsolved problem since the inception of machine learning, and becomes even more crucial in training the millions of parameters in neural networks.  The past work has investigated techniques for hyper-parameter tuning and search strategies, such as Bergstra, et. al., concluding that random search outperforms grid search \cite{bergstra2011algorithms}.  Likewise, the authors in \cite{shahriari2016taking}, \cite{snoek2012practical} take a Bayesian approach toward parameter estimation and optimization.  However, these efforts apply their strategies on image classfication tasks.

In relation to NMT, Britz, et. al. massively analyze neural network architectures and its variants~\cite{britz2017massive}.  Their approach incorporates a 2-layer bidirectional encoder/decoder with a multiplicative attention mechanism as a baseline architecture, with a 512-unit GRU and a dropout of 0.2 probability.  Their model parameters remained fixed and the studies varied the architecture, including depth layer, unidirectional vs bidirectional encoder/decoder, attention mechanism size, and beam search strategies. % and found that the depth of 2 performed best, and any deeper the model diverges.  
Likewise, Bahar et. al. compare various optimization strategies for NMT by switching to a different optimizer after $10k$ iterations, and found that Adam combined with other optimizers, such as SGD or annealing, increased the BLEU score by 2.4~\cite{bahar2017empirical}.  However, these approaches study a standard NMT system.  In addition, Wu, et. al. \cite{wu2016google} utilized the combination of Adam and SGD, where Adam ran for a fixed number of iterations with a 0.0002 learning rate, and switched to SGD with a 0.5 learning decay rate to slow down training, but did not perform hyper-parameter optimization.

To the best of our knowledge, there has not been any work comparing different hyper-parameter optimization strategies for NMT.  Moreover, our optimization strategies are demonstrated on a production-ready NMT system and explores parameter selection tradeoffs, in terms of performance and stability.

\section{Background}
Machine translation involves model design and model training.  In general, learning algorithms are viewed as a combination of selecting a \textit{model criterion}, defined as a family of functions, \textit{training}, defined as parameterization, and a procedure for appropriately optimizing this criterion.  The next subsections discuss how sentences are represented with a neural network and the optimization objectives used for training a model for a translation system.

\begin{figure}
\center
\includegraphics[scale=0.3]{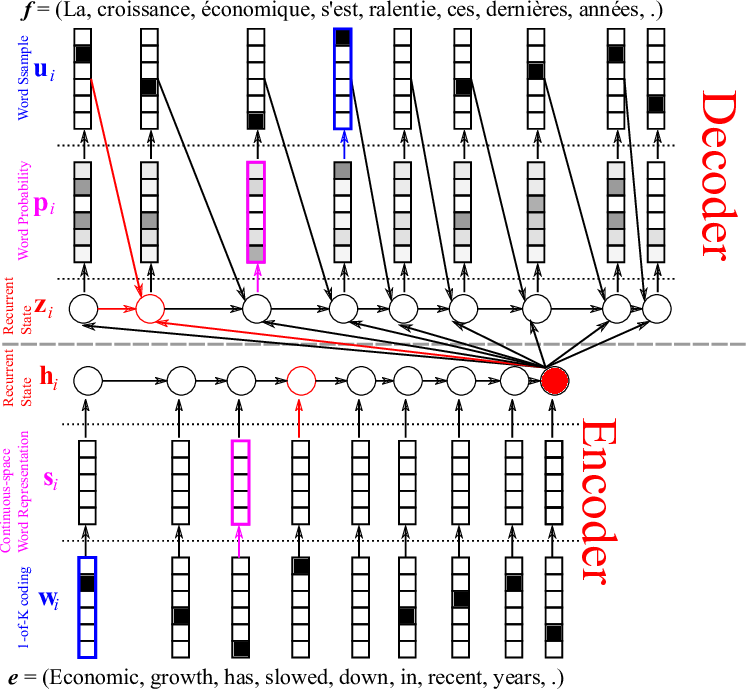}

\caption{RNN encoder-decoder, illustrating a sentence translation from English to French. The architecture includes a word embedding space, a 1-of-K coding and a recurrent state on both ends.\protect\footnotemark}
\label{fig:rnn}
\end{figure}

\footnotetext{\scriptsize{\url{https://devblogs.nvidia.com/introduction-neural-machine-translation-gpus-part-2/}}}

\subsection{Machine Translation}
This subsection discusses how neural networks can model language translation from a source to a target sequence.

\subsubsection{Recurrent Neural Networks}\label{sec:rnn}
Recurrent neural networks (RNN) are typically employed for neural machine translation because of its ability to handle variable length sequences.  RNNs capture unbounded context dependencies typical in natural language comprehension and speech recognition systems.  

For inputs $x_t$ and $y_t$, connection weight matrices $\mathbf{W}_{ih}$, $\mathbf{W}_{hh}$, $\mathbf{W}_{ho}$, indicating input-to-hidden, hidden-to-hidden and hidden-to-output, respectively, and activation function $f$, the recurrent neural network can be described as follows:
\begin{align}
h_t &= f_H (\mathbf{W}_{ih} x_t + \mathbf{W}_{hh} h_{t-1}) \\
y_t &= f_O (\mathbf{W}_{ho} h_t). 
\end{align}

RNNs learn a probability distribution over a sequence by being trained to predict the next symbol in a sequence.  The output at each timestep $t$ is the conditional probability distribution $p(x_t | x_{t-1}, ..., x_1)$.

\begin{table*}[t] % \small
   \caption{Stochastic gradient descent and its variants.}
  \label{tab:sgd_op}
  \centering
  \begin{tabular}{|p{2cm}|p{5cm}|p{5cm}|}
	\hline 
	\multicolumn{1}{|p{0.1\columnwidth}|}{\centering Optimizer} &
	\multicolumn{1}{p{0.1\columnwidth}|}{\centering Operations} &
	\multicolumn{1}{p{0.1\columnwidth}|}{\centering Description}  \\
	\hline\hline
        \texttt{SGD}  & $\begin{array}{l}
\mathbf{g}_t \leftarrow \bigtriangledown_{{\mathbf{\theta}}_t} J(\theta_t) \\
\mathbf{\theta}_{t+1} \leftarrow \mathbf{\theta}_t - \eta \mathbf{g}_t 
        \end{array}$ & $\mathbf{g}_t$ - gradient cost function, $\eta$ - learning rate, $\theta$ parameters
        \\ \hline
\texttt{AdaGrad} &  $\begin{array}{l}
\mathbf{g}_t \leftarrow \bigtriangledown_{{\mathbf{\theta}}_t} J(\theta_t)\\
\mathbf{\eta}_t \leftarrow \mathbf{\eta}_{t-1} + \mathbf{g}_t^2\\
\mathbf{\theta}_{t+1} \leftarrow \mathbf{\theta}_t - \frac{\eta}{\sqrt{\eta_t} + \epsilon}  \mathbf{g}_t 
\end{array}$ &  Divides $\eta$ by previous gradients, handles sparse data well 
\\ \hline
     \texttt{Adam} &  $\begin{array}{l}
\mathbf{g}_t \leftarrow \bigtriangledown_{{\mathbf{\theta}}_t} J(\theta_t)\\
\mathbf{\eta}_t \leftarrow \gamma \mathbf{\eta}_{t-1} + (1-\gamma) \mathbf{g}_t^2\\
\mathbf{{\hat{\eta}}} \leftarrow  \frac{\eta_t}{1-\gamma^t} \\
\mathbf{m}_t \leftarrow \mu \mathbf{m}_{t-1} + (1-\mu) \mathbf{g}_t\\
\mathbf{\hat{m}} \leftarrow \frac{\mathbf{m}_t}{1-\mu^t}\\
\mathbf{\theta}_{t+1} \leftarrow \mathbf{\theta}_t - \frac{\eta}{\sqrt{\hat{\eta}_t} + \epsilon}  \mathbf{\hat{m}}_t
     \end{array}$ &  $\mathbf{m}_t$ - decay mean of past gradients, $\hat{m}_t, \hat{n}_t$ - biased corrected terms that avoids zero initialization, $\gamma=0.9$, $\mu = 0.999$, $\epsilon = 10^8$
     \\ \hline

  \end{tabular}
\end{table*}

\begin{table*}[t] % \small
   \caption{Activation units for RNN.}
  \label{tab:act_rnn}
  \centering
  \begin{tabular}{|p{2cm}|p{5.8cm}|p{4cm}|}
	\hline 
	\multicolumn{1}{|p{0.1\columnwidth}|}{\centering Activation} &
	\multicolumn{1}{p{0.1\columnwidth}|}{\centering Operations} &
	\multicolumn{1}{p{0.1\columnwidth}|}{\centering Description}  \\
	\hline\hline
        \texttt{tanh}  & $\begin{array}{l}
\mathbf{s}_t \leftarrow {(e^{\mathbf{x}} - e^{-\mathbf{x}}})/{(e^{\mathbf{x}} + e^{-\mathbf{x}})}
        \end{array}$  & 
        hyperbolic tangent \\ \hline        	
        \texttt{LSTM}  & $\begin{array}{l}
\mathbf{i} \leftarrow \sigma(\mathbf{x}_t \mathbf{U}^i + \mathbf{s}_{t-1} \mathbf{W}^i)\\
\mathbf{f} \leftarrow \sigma(\mathbf{x}_t \mathbf{U}^f + \mathbf{s}_{t-1} \mathbf{W}^f) \\
\mathbf{o} \leftarrow \sigma(\mathbf{x}_t \mathbf{U}^o + \mathbf{s}_{t-1} \mathbf{W}^o) \\
\mathbf{g} \leftarrow \text{tanh}(\mathbf{x}_t \mathbf{U}^g + \mathbf{s}_{t-1} \mathbf{W}^g)\\
\mathbf{c}_t \leftarrow \mathbf{c}_{t-1} \circ \mathbf{f} + \mathbf{g} \circ \mathbf{i} \\
\mathbf{s}_t \leftarrow \text{tanh}(\mathbf{c}_t) \circ \mathbf{o}
        \end{array}$  & 
        3 gates, $c$ - internal memory, $o$ output, 2 tanh       
        \\ \hline
\texttt{GRU} &  $\begin{array}{l}
\mathbf{z} \leftarrow \sigma(\mathbf{x}_t \mathbf{U}^z + \mathbf{s}_{t-1} \mathbf{W}^z) \\
\mathbf{r} \leftarrow \sigma(\mathbf{x}_t \mathbf{U}^r + \mathbf{s}_{t-1} \mathbf{W}^r) \\
\mathbf{h} \leftarrow \text{tanh}(\mathbf{x}_t \mathbf{U}^h + (\mathbf{s}_{t-1} \circ \mathbf{r}) \mathbf{W}^h) \\
\mathbf{s}_t \leftarrow (1-z) \circ \mathbf{h} + \mathbf{z} \circ \mathbf{s}_{t-1}
\end{array}$   & 
2 gates, no internal memory, no output gates, 1 tanh
\\ \hline
  \end{tabular}
\end{table*}

\subsubsection{RNN Encoder-Decoder}
A RNN encoder-decoder (pictured in Fig.~\ref{fig:rnn}) encodes a variable-length sequence into a fixed vector representation, and decodes the fixed vector representation into a variable-length sequence \cite{cho2014learning}.  
%\begin{mydef}
The RNN encoder-decoder are separate neural networks that are jointly trained to maximize the conditional log-likelihood, defined as
\begin{align}
\label{eq:ll}
\argmax_{\theta} \frac{1}{N} \sum_{n=1}^{N} \text{log} \; p_{\theta} (t_n | s_n),
\end{align}
where $\theta$ represents the set of model parameters, each $\mathbf{s}_n, \mathbf{t}_n$ is a pair of input and output sequences from a parallel text corpus training set, and the output of the decoder from the encoder is differentiable.  
%\end{mydef}
% Gradient-based algorithms are used to estimate the model parameters, discussed in Sec~\ref{sec:grad}.  
A trained RNN encoder-decoder can generate a target sequence given an input sequence. 

\subsubsection{Neural Machine Translation}

Neural machine translation is defined as maximizing the conditional probability, $\arg \max_t p (\mathbf{t} | \mathbf{s}) \propto p(\mathbf{s}|\mathbf{t}) p(\mathbf{t})$, for a source $\mathbf{s}$ and target $\mathbf{t}$ sequence, where $p(\mathbf{s}|\mathbf{t})$ represents the translation model, and $p(\mathbf{t})$ represents the language model~\cite{sutskever2014sequence}, \cite{bahdanau2014neural}.  

Taking the log linear of $p (\mathbf{t} | \mathbf{s})$ yields, 
\begin{align}
 \text{log} \; p(\mathbf{t|s}) &= \sum_{n=1}^{N} w_n t_n (\mathbf{t,s}) + \text{log} \lambda(\mathbf{s}),
\end{align}
where $t_n$ and $w_n$ are the $n^{th}$ feature and weight, and $\lambda(s)$ is a normalization constant.  
The BLEU score provides a measure for optimizing weights during training.

\subsection{Optimization Objectives}
\label{sec:grad}
The following subsections describe the tuning of hyper-parameters that affect the performance of training a NMT system.  In particular, this work focuses on the optimizers, activation functions, and dropout.

\subsubsection{SGD Optimizers}

Stochastic gradient descent (\textit{SGD}), commonly used to train neural networks, updates a set of parameters $\theta$, where $\eta$ is the learning rate and $\mathbf{g}_t$ represents the gradient cost function, $J(\cdot)$.  
\textit{Adagrad} is an adaptive-based gradient method, where $\eta$ is divided by the square of all previous gradients, $\mathbf{\eta}_t$, plus $\epsilon$, a smoothing term to avoid dividing by zero.  As a result, larger gradients have less frequent updates, whereas smaller gradients have more frequent updates.  Adagrad handles sparse data well and does not require manual tuning of $\eta$.  
Adaptive moment estimation (\textit{Adam}) accumulates the decaying mean of past gradients, $\mathbf{m}_t$, and the decaying average of past squared gradients, $\eta_t$, referred to as the first and second moments, respectively.  The moments, $\hat{m}_t, \hat{n}_t$ are biased corrected terms that avoids initializing to zero.  $\gamma$ is usually set to 0.9, with $\mu = 0.999$, and $\epsilon = 10^8$.  Table~\ref{tab:sgd_op} displays SGD, AdaGrad and Adam optimizers.

\begin{table*}[t] % \small
   \caption{Dropout versus a standard update function.}
  \label{tab:drop}
  \centering
  \begin{tabular}{|p{2cm}|p{5.8cm}|p{4cm}|}
	\hline 
	\multicolumn{1}{|p{0.1\columnwidth}|}{\centering Optimizer} &
	\multicolumn{1}{p{0.1\columnwidth}|}{\centering Operations} &
	\multicolumn{1}{p{0.1\columnwidth}|}{\centering Description}  \\
	\hline\hline
        \texttt{Update}  & $\begin{array}{l}
z_i^{(l+1)} \leftarrow \mathbf{w}_i^{(l+1)} \mathbf{y}^l + b_i^{(l+1)}\\
y_i^{(l+1)} \leftarrow f(z_i^{(l+1)})
        \end{array}$  & 
        Standard update        
        \\ \hline
\texttt{Dropout} &  $\begin{array}{l}
r_j^{(l)} \sim \text{Bernoulli}(p)\\
\hat{\mathbf{y}}^{(l)} \leftarrow \mathbf{r}^{(l)} * \mathbf{y}^{(l)} \\
z_i^{(l+1)} \leftarrow \mathbf{w}_i^{(l+1)} \hat{\mathbf{y}}^l + b_i^{(l+1)} \\
y_i^{(l+1)} \leftarrow f(z_i^{(l+1)})
\end{array}$   & 
$r$ - Bernoulli rv, $p$ - dropout param
\\ \hline
  \end{tabular}
\end{table*}

\subsubsection{Activation Functions}
\label{sec:act_fn}
Activation functions serve as logic gates for recurrent neural networks that computes the hidden states, and include the hyperbolic tangent, long short term memory (LSTM)~\cite{hochreiter1997long}, and gated recurrent unit (GRU)~\cite{cho2014learning}.  Table~\ref{tab:act_rnn} displays the hyperbolic tangent, LSTM and GRU activation functions.

To address the vanishing gradients problem associated with learning long-term dependencies in RNNs, LSTMs and GRUs employ a gating mechanism when computing the hidden states.  For \textit{LSTM}s, note that the input $i$, forget $f$ and hidden $h$ gates are the same equations except with different parameter matrices.    
$g$ is a hidden state, based on the current input and previous hidden state.  $c_t$ serves as the internal memory, which is a combination of the previous memory, $c_{t-1}$, multiplied by the input gate.   
The hidden state, $s_t$, is calculated by multiplying $c_t$ and the output gate.  On the other hand, a \textit{GRU} employs a reset gate $r$ and an update gate $u$.  The reset gate $r$ determines how to combine the new input with the previous memory, whereas the update gate $u$ defines how much of the previous memory to retain.  If the reset gates were set to 1's and the update gates to 0's, this would result in a vanilla RNN.  

The differences between the approaches to compute hidden units are that GRUs have 2 gates, whereas LSTMs have 3 gates.  GRUs do not have an internal memory and output gates, compared with LSTM which uses $c$ as its internal memory and $o$ as an output gate.  The GRU input and forget gates are coupled by an update gate $z$, and the reset gate $r$ is applied directly to the previous hidden state.  Also, GRUs do not have a 2nd non-linearity operation, compared to LSTMs, which uses two hyperbolic tangents.

\subsubsection{Dropout}

In a fully-connected, feed-forward neural network, \textit{dropout} randomly retains connections within hidden layers while discarding others \cite{srivastava2014dropout}.   Table~\ref{tab:drop} displays a standard hidden update function on the top, whereas a version that decides whether to retain a connection is displayed on the bottom.  $\hat{\mathbf{y}}^{(l)}$ is the thinned output layer, and retaining a network connection is decided by a Bernoulli random variable $r^{(l)}$ with probability $p(\cdot)=1$.

\subsection{Combination of Optimizers}
Since the learning trajectory significantly affects the training process, it is required to select and tune the proper types of hyper-parameters to yield good performance.   The construction of the RNN cell with activation functions, the optimizer and its learning rate, and the dropout rates all have an affect on how the training progresses, and whether good accuracy can be achieved.  

\begin{table} 
%\tiny
\caption{Datasets used in experiments.}
\centering 
\begin{tabular}{|r|cc|} \hline 
\ & RO$\rightarrow$EN, EN$\rightarrow$RO & DE$\rightarrow$EN, EN$\rightarrow$DE \\ \hline \hline
\ Train & corpus.bpe (2603030) & corpus.bpe (4497879) \\ %\hline
\ Valid & newsdev2016.bpe (1999) & newstest2014.bpe (3003) \\ %\hline
Test & newstest2016.bpe (1999) & newstest2016.bpe (2999)  \\ \hline
\end{tabular}
\label{tab:data}
\end{table}

\begin{table}
\caption{Graphical processors used in this experiment.}
\centering
\begin{tabular}{|r|cc|} \hline
\ &P100 & V100\\ \hline \hline
\ CUDA capability  & 6.0 & 7.0 \\ %\hline
\ Global memory (MB) & 16276 & 16152 \\ %\hline
\ Multiprocessors (MP) & 56 & 80 \\ %\hline
\ CUDA cores per MP & 64 & 64 \\ %\hline
\ CUDA cores & 3584 & 5120 \\ %\hline
\ GPU clock rate (MHz) & 405  & 1380 \\ %\hline
\ Memory clock rate (MHz) & 715 & 877 \\ %\hline
L2 cache size (MB) & 4.194 & 6.291 \\ %\hline
Constant memory (bytes) & 65536 & 65536 \\ %\hline
Shared mem blk (bytes) & 49152 & 49152 \\ %\hline
Registers per block & 65536 & 65536 \\ %\hline
Warp size & 32 & 32 \\ %\hline
Max threads per MP & 2048 & 2048 \\ %\hline
Max threads per block & 1024 & 1024 \\ 
CPU (Intel) & Ivy Bridge & Haswell \\ \hline\hline
Architecture family &  Pascal & Volta \\
\hline\end{tabular}
\label{tab:gpu}
\end{table}

\begin{table}[t]
    \caption{Hardware and execution environment information.}
  \centering
  \begin{tabular}{|r|M{0.3\columnwidth}|M{0.3\columnwidth}|}
	\hline
%	\multicolumn{1}{|p{0.08\columnwidth}|}{\centering\tabhead{Setting}} &
%	\multicolumn{1}{p{0.15\columnwidth}|}{\centering\tabhead{Value}}\\
%	\hline\hline
	\text{Architecture} & Haswell & Ivy Bridge\\ \hline \hline
	\text{Model}& E5-2698 v3 & Xeon X5650\\
	%\hline
	\text{Clock speed} & 2.30 GHz & 2.67 GHz\\
    \text{Node count} & 4, 14  & 6\\
    \text{GPUs} & 4 $\times$ V100  & 4 $\times$ P100\\
	%\hline
	\text{Memory}& 256 GB & 50 GB\\
	\text{Linux kernel} & 3.10.0-229.14.1 & 2.6.32-642.4.2\\ \hline
	\text{Compiler}& \multicolumn{2}{|M{0.65\columnwidth}|} {CUDA v9.0.67 }\\	
	\text{Flags} & \multicolumn{2}{|M{0.65\columnwidth}|}{\{$\mathtt{`g'}$, $\mathtt{`lineinfo'}$,\, {$\mathtt{`arch=sm\_cc'}$\}} }\\
	\hline
  \end{tabular}
  \label{tab:hw}
\end{table}

\section{Marian NMT}
Marian~\cite{junczys2018marian} is an efficient NMT framework written in C++, with support for multi-node and multi-GPU training and CPU/GPU translation capabilities.  Marian is currently being developed and deployed by the Microsoft Translator team.  Table~\ref{tab:marian-params} displays parameters involved with tuning a neural machine translation system, categorized by model, training and validation, with values and types in brackets, and its default value, if any.  The types of models in Marian include RNNs and Transformers~\cite{vaswani2017attention}.

The translation system evaluated in this study is a sequence-to-sequence model with single layer RNNs for both the encoder and decoder.  The RNN in the encoder is bi-directional and the decoder is sequence-to-sequence.  Depth, also referred to as \textit{deep transitions}~\cite{koehn2017neural}, is achieved by stacking activation blocks, resulting in tall RNN cells for every recurrent step.  The encoder consists of four activation blocks per cell, whereas the decoder consists of eight activation blocks, with an attention mechanism placed between the first and second block.  Word embedding sizes were set at 512, the RNN state size was set to 1024, and layer normalization was applied inside the activation blocks and the attention mechanism.

\begin{table*} 
%\small
\caption{Marian hyper-parameters, with options in brackets.}
\centering 
\begin{tabular}{|p{0.4cm}|p{3cm}|p{9cm}|}  \hline 
  \  & dimensions & vocab [vect]   \\ %\hline
    \  &  & embed [int]    \\ %\cline{2-3}
   \    & RNN & dim [int]   \\ 
     \  &  & type [bi-dir, bi-unidir, s2s]   \\ %\hline
     \   &  & cell: type [gru, lstm, tanh], depth [1, 2, ...], transition cells [1, 2, ...]  \\ %\hline     
     \  &  & skip [bool]   \\ %\hline
     \ \parbox[t]{1mm}{\multirow{2}{*}{\rotatebox[origin=c]{90}{Model}}}   &  & layer norm [bool]   \\ %\hline
     \ &  & tied embeddings [src, trg, all]    \\ %\hline
     \  &  & dropout [float]    \\% \cline{2-3} 
     \  & transformer & heads [int]    \\ %\hline  
     \  &  & no projection    \\ %\hline                
     \  &  & tied layers [vector]    \\ %\hline  
     \  &  & guided alignment layer    \\ %\hline           
     \  &  & preproc, postproc, post-emb [dr, add, norm]    \\ %\hline                
     \  &  & dropout [float]    \\ %\hline                
   \hline \hline
 \ & cost & [ce-mean, ce-mean, words, ce-sum, perplexity]   \\ %\hline
 \ & after-epochs & $[ \infty ]$     \\ %\hline
 \  & max length & [int=50]   \\ %\hline
 \ & system & GPUs, threads   \\ % \hline
 \ & mini-batch & size, words, fit, fit-step [int, int, bool, uint]  \\ 
 \ & optimizer & [sgd, adgrad, adam]  \\ 
 \ \parbox[t]{1mm}{\multirow{2}{*}{\rotatebox[origin=c]{90}{Training}}}  & learn rate & decay: strategy [epochs, stalled, epoch + batches, ep+stalled], start, frequency, repeat warmup, inverse sqrt, warmup \\ 
 \  & label smoothing & [bool]   \\
 \ & clip norm & [float=1]  \\ 
 \ & exponential smoothing & [float=0]   \\ 
 \ & guided alignment & cost [ce, mean, mult], weight [float=0.1]   \\ 
 \ & data weighting  & type [sentence, word]  \\ 
 \ & embedding & vectors, norm, fix-src, fix-trg  \\ \hline \hline
\ & frequency &    \\ %\hline
    \  & metrics & [ce, ce-words, perplexity, valid-script, translation, bleu, bleu-detok]    \\ %\cline{2-3}
   \  \parbox[t]{1mm}{\multirow{2}{*}{\rotatebox[origin=c]{90}{Validation}}}  & early stopping & [int=10]   \\ 
     \  & beam size & [int=12]   \\ %\hline
     \   & normalize & [float=0]  \\ %\hline     
     \  & max-length-factor &  [float=3]   \\ %\hline
     \  & word penalty & [float]   \\ %\hline
     \  & mini-batch & [int=32]    \\ 
     \  & max length & [int=1000]   \\ %\hline     
     \hline 
\end{tabular}
\label{tab:marian-params}
\end{table*}

\begin{table*} 
%\tiny
\caption{BLEU scores for validation (top) and test (bottom) datasets.}
\centering 
\begin{tabular}{|c|r|cccc|cccc|} \hline 
\ & & \multicolumn{2}{c}{ro$\rightarrow$en}  & \multicolumn{2}{c|}{en$\rightarrow$ro} & \multicolumn{2}{c}{de$\rightarrow$en}  & \multicolumn{2}{c|}{en$\rightarrow$de} \\
\ cell & learn-rt & P100 & V100 & P100 & V100 & P100 & V100 & P100 & V100 \\ \hline \hline
 \ GRU & 1e-3 & \cellcolor{gray!25}35.53 & 35.43 & 19.19 & 19.28 & 28.00 & 27.84 & 20.43 & 20.61 \\ %\hline
\ & 5e-3 & 34.37 & 34.05 & 19.07 & 19.16 & 26.05 & 22.16 & n/a & 19.01\\ %\hline
\ & 1e-4 & 35.47 & 35.46 & 19.45 & 19.49 & 27.37 & 27.81 & dnf & 21.41\\ \hline
\ LSTM & 1e-3 & 34.27 & \cellcolor{gray!25}35.61 & 19.29 & \cellcolor{gray!25}19.64 & \cellcolor{gray!25}28.62 & \cellcolor{gray!25}28.83 & \cellcolor{gray!25}21.70 & \cellcolor{gray!25}21.69 \\
\ & 5e-3 & 35.05 & 34.99 & \cellcolor{gray!25} 19.48 & 19.43 & n/a & 24.36 & 18.53 & 18.01 \\
\ & 1e-4 & 35.41 & 35.28 & 19.43 & 19.48 & n/a & 28.50 & dnf & dnf \\ \hline \hline
 \ GRU & 1e-3 & \cellcolor{gray!25}34.22 & \cellcolor{gray!25}34.17 & 19.42 & 19.43 & 33.03 & 32.55 & 26.55 & 26.85 \\ %\hline
\ & 5e-3 & 33.13 & 32.74 & 19.31 & 18.97 & 31.04 & 26.76 & n/a & 26.02\\ %\hline
\ & 1e-4 & 33.67 & 34.44 & 18.98 & \cellcolor{gray!25}19.69 & \cellcolor{gray!25}33.15 & 33.12 & dnf & 28.43\\ \hline
\ LSTM & 1e-3 & 33.10 &  33.95 & \cellcolor{gray!25}19.56 & 19.08 & 33.10 & \cellcolor{gray!25}33.89 & \cellcolor{gray!25}28.79 & \cellcolor{gray!25}28.84 \\
\ & 5e-3 & 33.10 & 33.52 & 19.13 & 19.51 & n/a & 29.16 & 24.12 & 24.12 \\
\ & 1e-4 & 33.29 & 32.92 & 19.14 & 19.23 & n/a & 33.44 & dnf & dnf \\ \hline

\end{tabular}
\label{tab:bleu}
\end{table*}

\begin{table*} 
%\tiny
\caption{Dropout rates, BLEU scores and total training time for test set, comparing systems.}
\centering 
\begin{tabular}{|c|r|cccc|cccc|} \hline 
\ & & \multicolumn{4}{c|}{ro$\rightarrow$en}  & \multicolumn{4}{c|}{de$\rightarrow$en} \\
\ cell & dropout & P100 & $t$ & V100 & $t$ & P100 & $t$ & V100 & $t$ \\ \hline \hline
 \ GRU & 0.0 & 34.47 & 6:29 & 34.47 & 4:43 & 32.29 & 9:48 & 31.61 & 6:15  \\ %\hline
\ & 0.2 & \cellcolor{gray!25}35.53 & \cellcolor{gray!25}8:48 & 35.43 & 6:21 & \cellcolor{gray!25}33.03 & \cellcolor{gray!25}18:47 & \cellcolor{gray!25}32.55 & \cellcolor{gray!25}19:40 \\ %\hline
\ & 0.3 & 35.36 & 12.21 & \cellcolor{gray!25}35.15 & \cellcolor{gray!25}7:28 & 31.36 & 10:14 & 31.50 & 9:33 \\ 
\ & 0.5 & 34.50 & 12:20 & 34.67 & 17:18 & 29.64 & 11:09 & 30.21 & 11:09 \\ \hline
\ LSTM & 0.0 & 34.84 & 6:29 & 34.65 & 4:46 & 32.84 & 12:17 & 32.88 & 7:37 \\
\ & 0.2 & 34.27 & 8:10 & \cellcolor{gray!25}35.61 & \cellcolor{gray!25}6:34 & 33.10 & 16:33 & \cellcolor{gray!25}33.89 & \cellcolor{gray!25}13:39 \\
\ & 0.3 & \cellcolor{gray!25}35.67 & \cellcolor{gray!25}9:56 & 35.37 & 11:29 & \cellcolor{gray!25}33.45 & \cellcolor{gray!25}20.02 & 33.51 & 15:51 \\
\ & 0.5 & 34.50 & 15:13 & 34.33 & 12:45 & 32.67 & 20:02 & 32.20 & 13:03 \\  \hline
\end{tabular}
\label{tab:dropout}
\end{table*}

\section{Experiments}
The experiments were carried out on the WMT 2016 \cite{junczys2016log} translation tasks for the Romanian and German languages in four directions:  EN $\rightarrow$ RO, RO $\rightarrow$ EN,  EN $\rightarrow$ DE, and DE $\rightarrow$ EN.  The datasets and its characteristics used in the experiments are listed in Table~\ref{tab:data}, with number of sentence examples in parenthesis.  Table~\ref{tab:data} shows that for WMT 2016 EN $\rightarrow$ RO and RO $\rightarrow$ EN, the training data consisted of 2.6M English and Romanian sentence pairs, whereas for WMT 2016 EN $\rightarrow$ DE and DE $\rightarrow$ EN, the training corpus consisted of approximately 4.5M German and English sentence pairs.  Validation was performed on 1000 sentences of the $\mathtt{newsdev2016}$ corpus for RO, and on the $\mathtt{newstest2014}$ corpus for DE.   %We stopped training after 200K mini-batch iterations, and evaluated the models every 5K.  
The $\mathtt{newstest2016}$ corpus consisted of 1999 sentences for RO and 2999 sentences for DE, and was used as the test set.  We evaluated and saved the models every 10K iterations and stopped training after 500K iterations.

All experiments used bilingual data without additional monolingual data.  
We used the joint byte precision encoding (BPE) approach \cite{sennrich2015neural} in both the source and target sets, which converts words to a sequence of subwords.  For all four tasks, the number of joint-BPE operations were 20K.  All words were projected on a 512-dimensional embedding space, with vocabulary dimensions of $66000 \times 50000$.  The mini-batch size was determined automatically based on the sentence length that was able to fit in GPU global memory, set at 13000 MB for each GPU.  .

Beam search was used for decoding, with the beam size set to 12.  The translation portion consisted of recasing and detokenizing the translated BPE chunks.  The trained models compared different hyper-parameter strategies, including the type of optimizer, the activation function, and the amount of dropout applied.  The number of parameters were initialized with the same random seed.   
The systems were evaluated using the case-sensitive BLEU score computed by Moses SMT~\cite{koehn2007moses}.

We compared models trained on two different types of GPUs (P100 Pascal, V100 Volta), listed on Table~\ref{tab:gpu}.  The corresponding CPUs are listed on Table~\ref{tab:hw}.  Each ran with four GPUs.  The dataset was partitioned across 4 GPUs, and a copy of the model was executed on each GPU.

\begin{figure*}
\centering
\includegraphics[width=1.6\columnwidth]{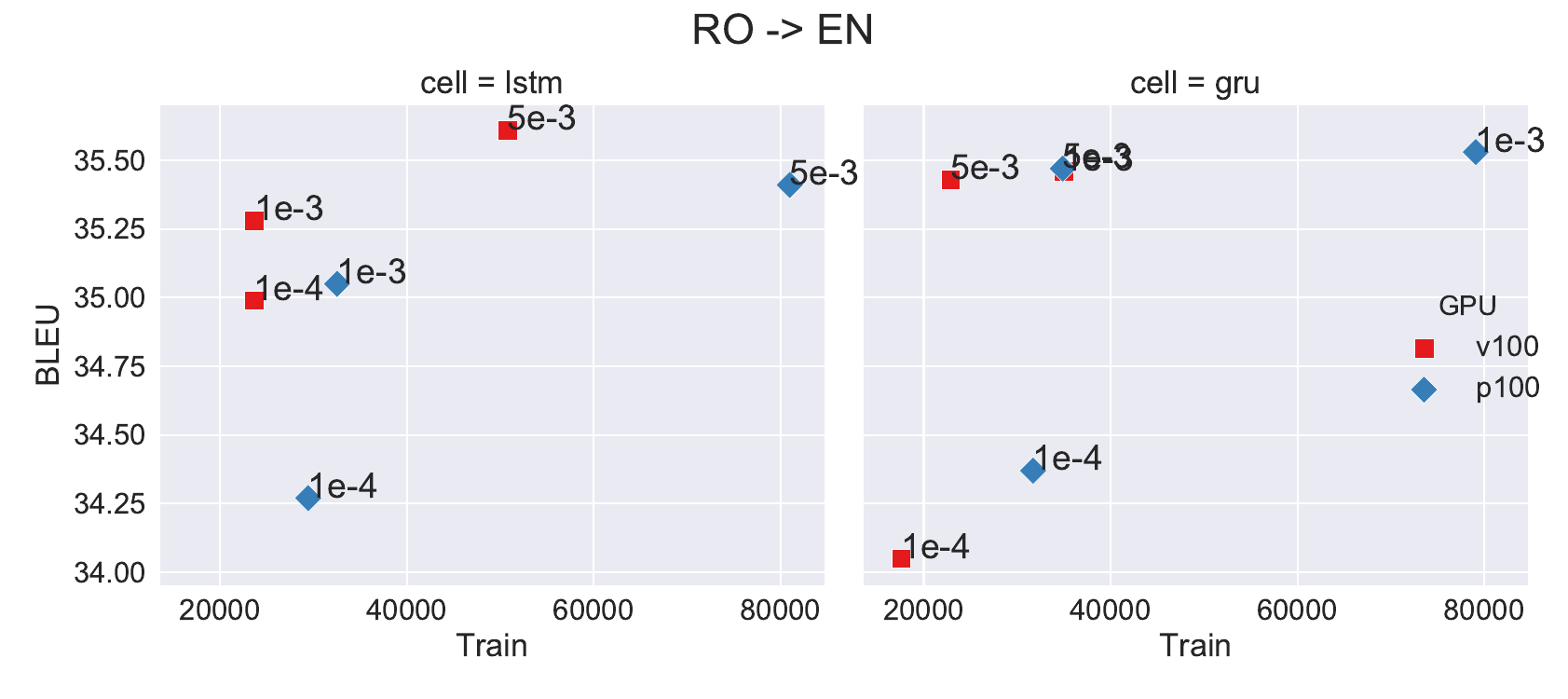}
\includegraphics[width=1.6\columnwidth]{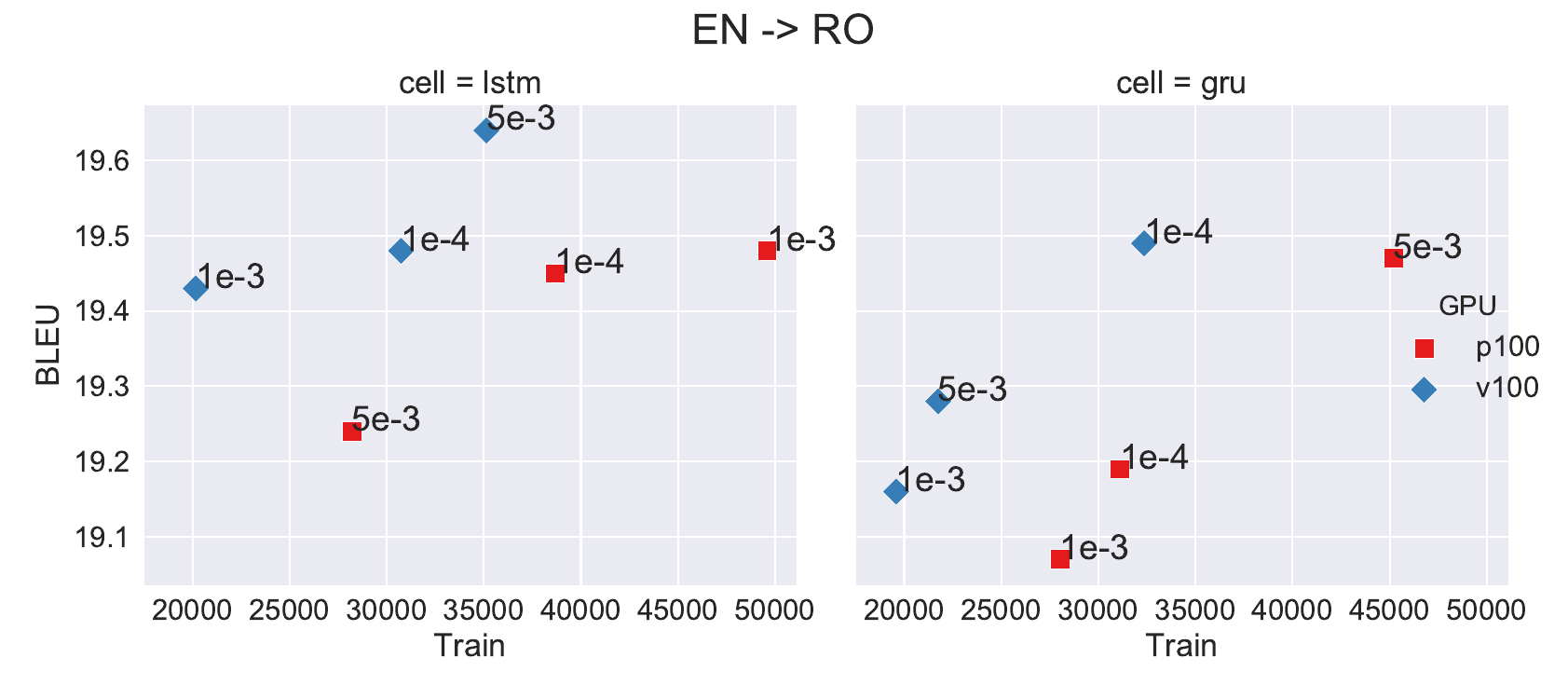}
\includegraphics[width=1.5\columnwidth]{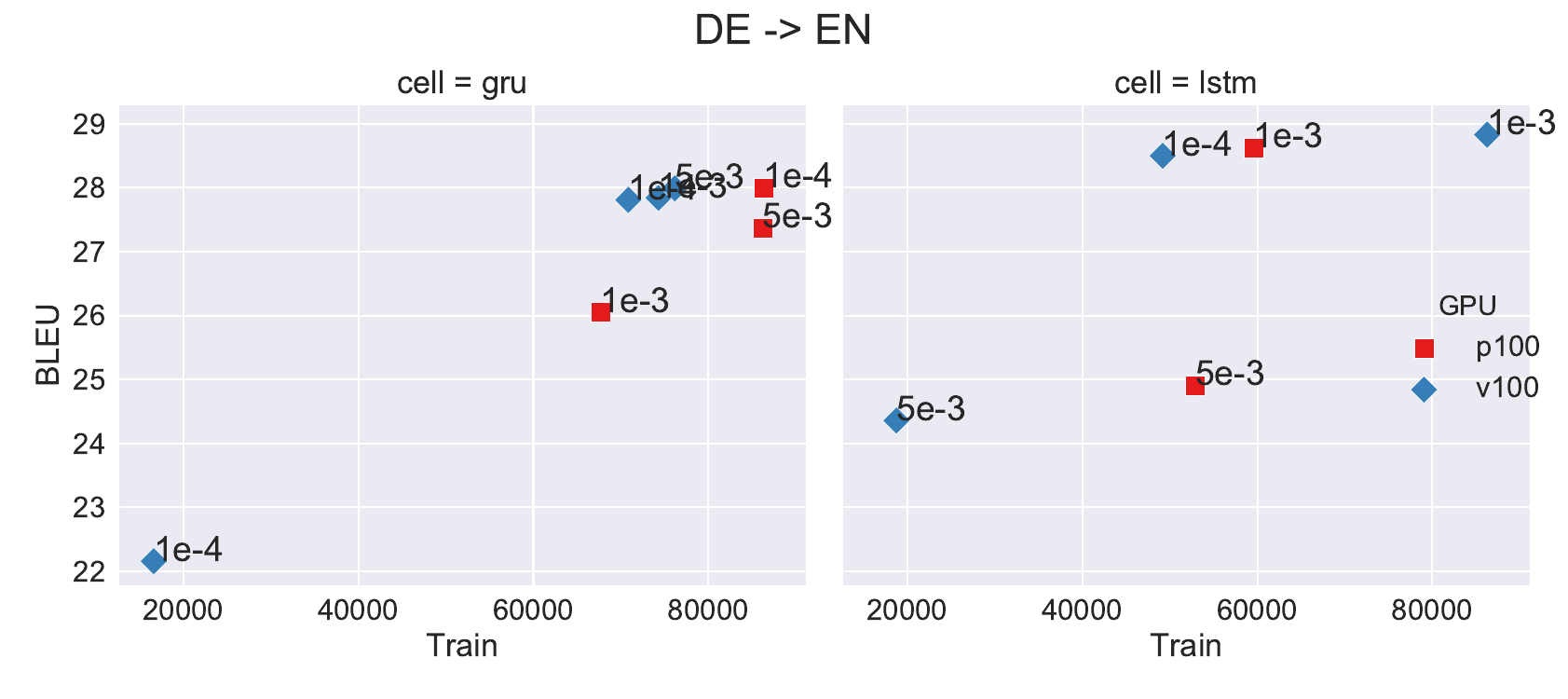}
\includegraphics[width=1.5\columnwidth]{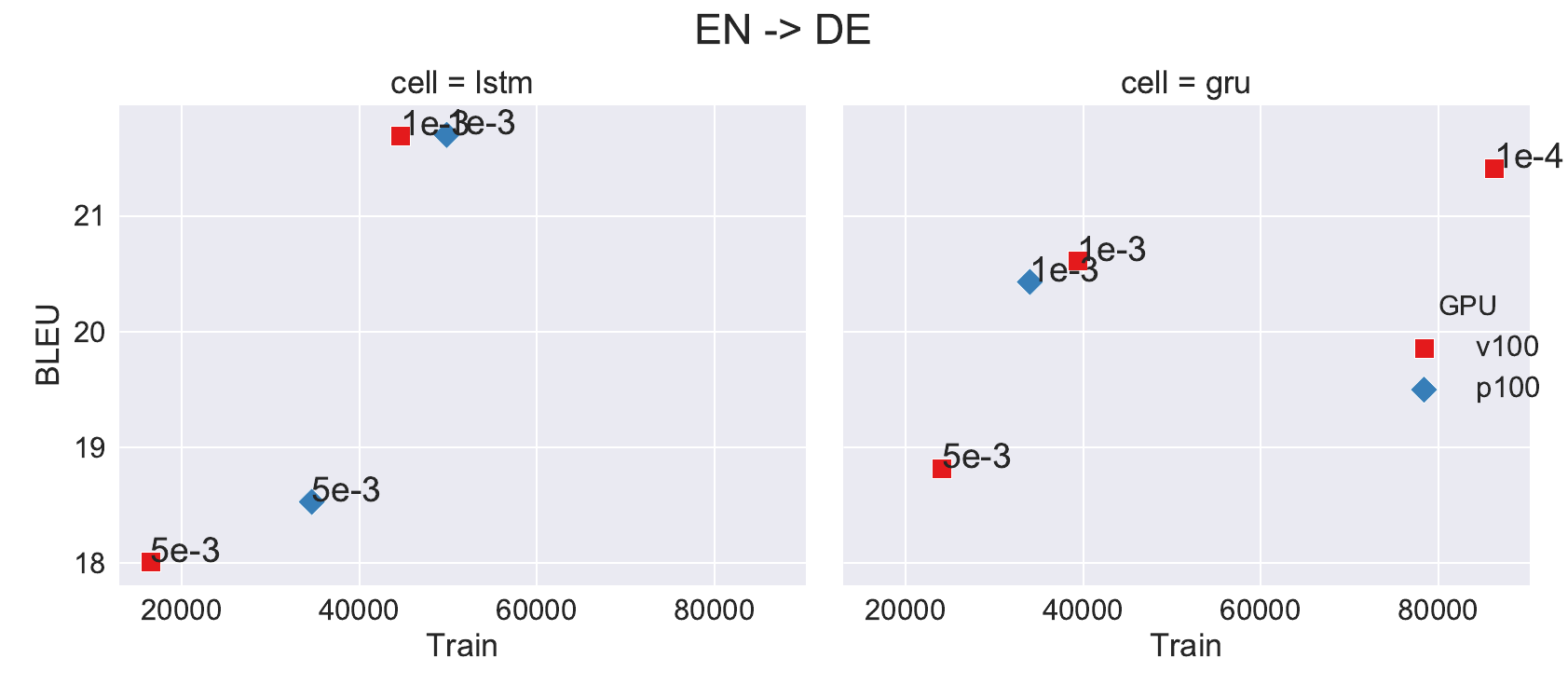}
\caption{BLEU scores as a function of training time (seconds), comparing GPUs (color), activation units (sub-columns), learning rates and translation directions.}
\label{fig:bleu_time}
\end{figure*}

\section{Analysis}

This section analyzes the results of the evaluated NMT systems in terms of translation quality, training stability, convergence speed and tuning cost.

\begin{figure*}
\centering
\includegraphics[width=1.7\columnwidth]{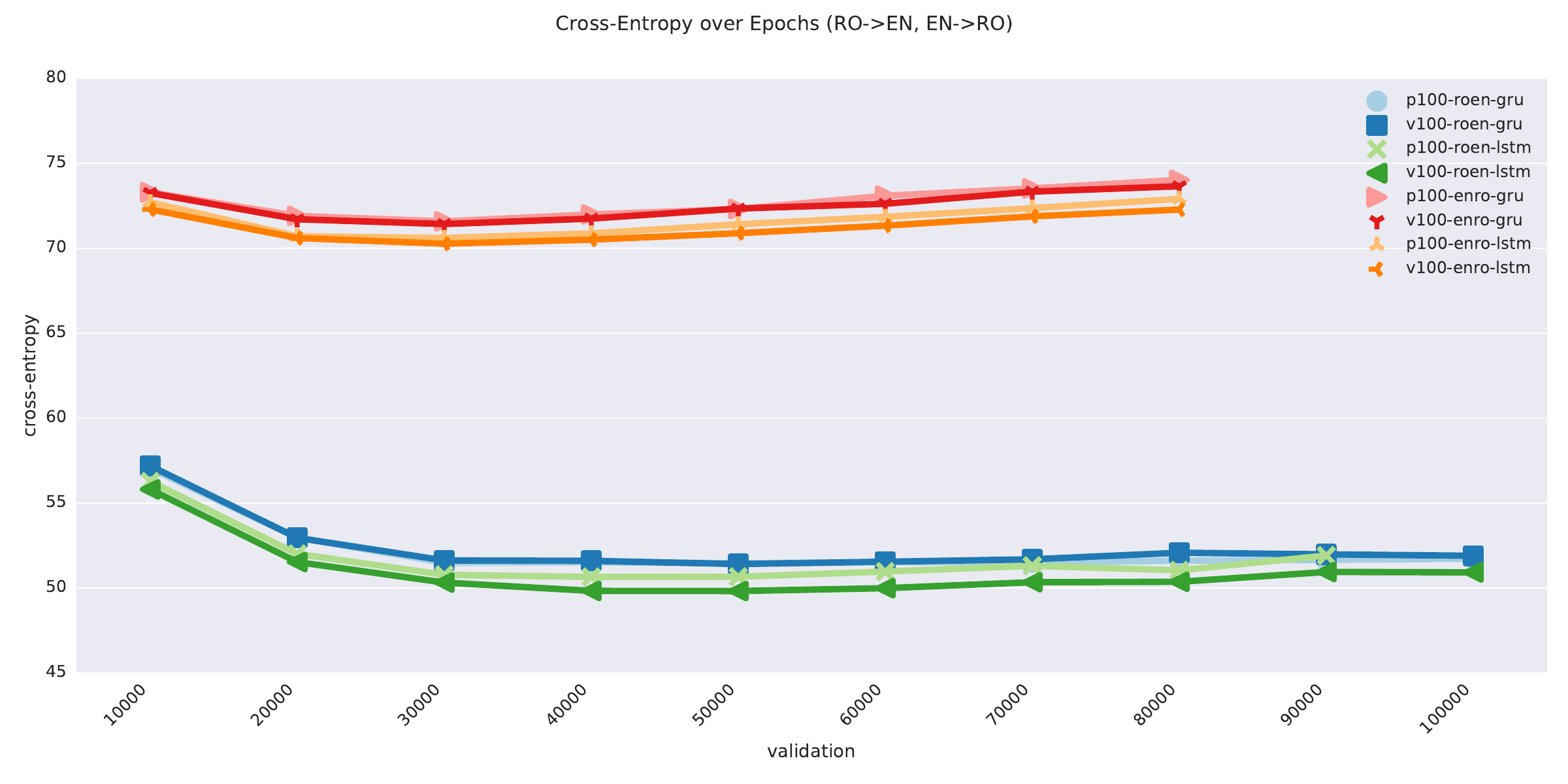}
\caption{Cross entropy over the number of epochs for RO $\rightarrow$ EN and EN $\rightarrow$ RO, comparing activation functions and GPUs.}
\label{fig:ce-ro}
\end{figure*}

\subsection{Translation Quality}

Table~\ref{tab:bleu} shows BLEU scores calculated for four translation directions for the validation sets (top) and the test sets (bottom), comparing learning rates, activation functions and GPUs.  Note that entries with $\mathtt{n/a}$ means that no results were available, whereas entries with $\mathtt{dnf}$ indicates training time that did not complete within 24 hours.  For the validation sets, LSTMs were able to achieve higher accuracy rates, whereas in the test set GRUs and LSTMs were about the same.  Also, note that the best performing learning rates were usually at a lower value (e.g. 1e-3).  The type of hidden unit mechanism (e.g LSTM vs GRU) and the learning rate can affect the overall accuracy achieved, as demonstrated by Table~\ref{tab:bleu}.

%\begin{figure*}
%\centering
%\includegraphics[width=1.4\columnwidth]{img/bleu-time-roen}
%\includegraphics[width=1.4\columnwidth]{img/bleu-time-enro}
%%\includegraphics[width=1\columnwidth]{img/bleu-time-deen}
%%\includegraphics[width=1\columnwidth]{img/bleu-time-ende}
%\caption{BLEU scores as a function of training time (seconds), comparing GPUs (color), activation units (sub-columns), learning rates and translation directions.}
%\label{fig:bleu_time}
%\end{figure*}

\begin{figure*}
\centering
\includegraphics[width=1.8\columnwidth]{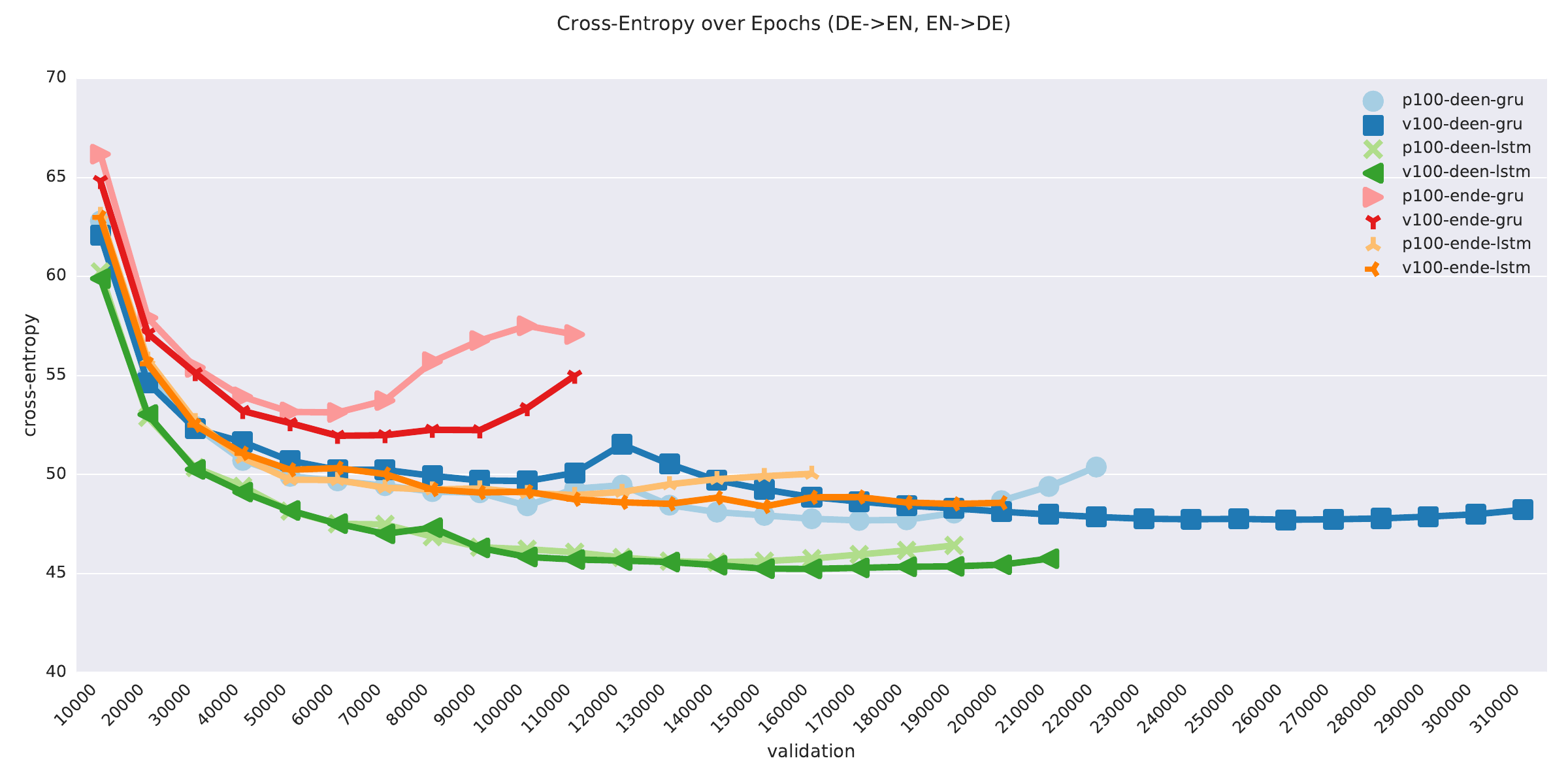}
\caption{Cross-entropy over the number of epochs for DE $\rightarrow$ EN and EN $\rightarrow$ DE, comparing activation functions and GPUs.}
\label{fig:ce-de}
\end{figure*}

Table~\ref{tab:dropout} displays various dropout rates applied for translation directions RO $\rightarrow$ EN and DE $\rightarrow$ EN, comparing hidden units, GPUs and overall training time.  The learning rate was evaluated at 0.001, the rate that achieved the highest BLEU score, as evident in Table~\ref{tab:bleu}.  Generally speaking, increasing the dropout rates also increased training time.  This may be the result of losing network connections when applying the dropout mechanism, but at the added benefit of avoiding overfitting.  This is evident in Table~\ref{tab:dropout}, where applying some form of dropout will result in a trained model achieving higher accuracies.   The best performance can be seen when the dropout rate was set at 0.2 to 0.3.  This confirms that some form of skip connection mechanism is necessary to prevent the overfitting of models under training.

Figure~\ref{fig:bleu_time} shows BLEU score results as a function of training time, comparing GPUs, activation units, learning rates and translation directions.  Note that in most cases a learning rate of 0.001 achieves the higher accuracy in most cases, at the cost of higher training time.  Also, note the correlation between longer training time and higher BLEU scores in most cases.  In some cases, the models were able to converge at a faster rate (e.g. Fig.~\ref{fig:bleu_time} upper left, RO$\rightarrow$EN, GRU with learning rate of 0.005 vs 0.001).

\begin{table*} 
% \small
\caption{Words-per-second (average) and number of epochs, comparing activation units, learning rates and GPUs.}
\centering 
\begin{tabular}{|c|r|cccc|cccc|} \hline 
\ & & \multicolumn{2}{c}{words-per-sec}  & \multicolumn{2}{c|}{validation} &
\multicolumn{2}{c}{words-per-sec}  & \multicolumn{2}{c|}{validation} \\
%\ & & \multicolumn{2}{c}{ro$\rightarrow$en}  & \multicolumn{2}{c|}{en$\rightarrow$ro} & \multicolumn{2}{c}{de$\rightarrow$en}  & \multicolumn{2}{c|}{en$\rightarrow$de} \\
\ cell & learn-rt & P100 & V100 & P100 & V100 & P100 & V100 & P100 & V100 \\ \hline \hline
\ & & \multicolumn{4}{c|}{ro$\rightarrow$en} & \multicolumn{4}{c|}{en$\rightarrow$ro}  \\ \hline
 \ GRU & 1e-3 & 33009.23 & \cellcolor{gray!25}45762.54 & 18000 &\cellcolor{gray!25}18000 & 29969.14 & \cellcolor{gray!25}42746.15 & 15000 & \cellcolor{gray!25}15000 \\ %\hline
\ & 5e-3 & 32965.23 & 24253.14 & 19000 & 8000 & 30223.89 & 23144.62 & 17000 & 10000\\ %\hline
\ & 1e-4 & 32828.61 & 24341.96 & 44000 & 16000 & 29959.34 & 23277.51 & 25000 & 14000 \\ \hline
\ LSTM & 1e-3 & 29412.87 & 40534.06 & 15000 & 16000 & 27282.54 & \cellcolor{gray!25}38131.13 & 14000 & \cellcolor{gray!25}14000 \\
\ & 5e-3 & 29536.65 & 40598.24 & 16000 & 16000 & 27245.42 & 37384.46 & 19000 & 21000 \\
\ & 1e-4 & 29478.51 & \cellcolor{gray!25}41441.37 & 40000 & \cellcolor{gray!25}35000 & 27002.60 & 38118.79 & 25000 & 25000 \\ \hline \hline

\ & & \multicolumn{4}{c|}{de$\rightarrow$en} & \multicolumn{4}{c|}{en$\rightarrow$de}  \\
% \ & & \multicolumn{2}{c}{words-per-sec (avg)}  & \multicolumn{2}{c|}{validation} &
% \multicolumn{2}{c}{words-per-sec (avg)}  & \multicolumn{2}{c|}{validation} \\
%\ & & \multicolumn{2}{c}{ro$\rightarrow$en}  & \multicolumn{2}{c|}{en$\rightarrow$ro} & \multicolumn{2}{c}{de$\rightarrow$en}  & \multicolumn{2}{c|}{en$\rightarrow$de} \\
%\ cell & learn-rt & P100 & V100 & P100 & V100 & P100 & V100 & P100 & V100 \\ \hline \hline
\hline
 \ GRU & 1e-3 & 28279.53 & \cellcolor{gray!25}38026.87 & 20000 & \cellcolor{gray!25}28000 & 28367.91 & \cellcolor{gray!25}39995.48 & 10000 & \cellcolor{gray!25}10000 \\ %\hline
\ & 5e-3 & 28215.40 & 19819.59 & 25000 & 4000 & n/a & 39944.10 & n/a & 16000\\ %\hline
\ & 1e-4 & 28367.54 & 33218.70 & 26000 & 32000 & dnf & 39993.89 & dnf & 36000 \\ \hline
\ LSTM & 1e-3 & 24995.64 & 33507.31 & 16000 & 17000 & 25245.67 & \cellcolor{gray!25}35122.54 & 13000 & \cellcolor{gray!25}17000 \\
\ & 5e-3 & 25210.15 & 33740.92 & 14000 & 7000 & 25049.21 & 33649.20 & 9000 & 6000 \\
\ & 1e-4 & dnf & \cellcolor{gray!25}34529.58 & dnf & \cellcolor{gray!25}31000 & dnf & dnf & dnf & dnf \\ \hline

\end{tabular}
\label{tab:wps_ro}
\end{table*}

\subsection{Training Stability}

Figure~\ref{fig:ce-ro} shows the cross-entropy scores for the RO $\rightarrow$ EN and EN $\rightarrow$ RO translation tasks, comparing different activation functions (GRU vs. LSTM), with learning rates at 0.001.  Note the training stability patterns that emerge from this plot, which is highly correlated with the translation direction.  The activation function (GRU vs LSTM) during validation also performed similarly across GPUs and was also highly correlated with the translation direction.  Cross-entropy scores for the EN $\rightarrow$ RO translation direction were more or less the same.  However, for RO $\rightarrow$ EN, a LSTM that executed on a P100 converged the earliest by one iteration.  

Figure~\ref{fig:ce-de} shows the same comparison of cross-entropy scores over epochs for DE $\rightarrow$ EN and EN $\rightarrow$ DE translation tasks.  Note that the behavior for this translation task was wildly different for all systems.  Not only did it take more epochs to converge compared to Fig~\ref{fig:ce-ro}, but also how well the system progressed also varied, as evident in the cross-entropy scores during validation.  When comparing hidden units, LSTMs outperformed GRUs in all cases.  When comparing GPUs, the V100 performed better than the P100 in terms of cross-entropy, but took longer to converge in some cases (e.g. v100-deen-lstm, v100-ende-lstm).  Also, note that the behavior of the translation task EN $\rightarrow$ DE for a GRU hidden unit never stabilized, as evident in both the high cross-entropy scores and the peaks toward the end.  The LSTM was able to achieve a better cross-entropy score overall, with nearly a 8 point difference for DE $\rightarrow$ EN, compared with the GRU.

\begin{figure*}
\centering
\includegraphics[width=1.7\columnwidth]{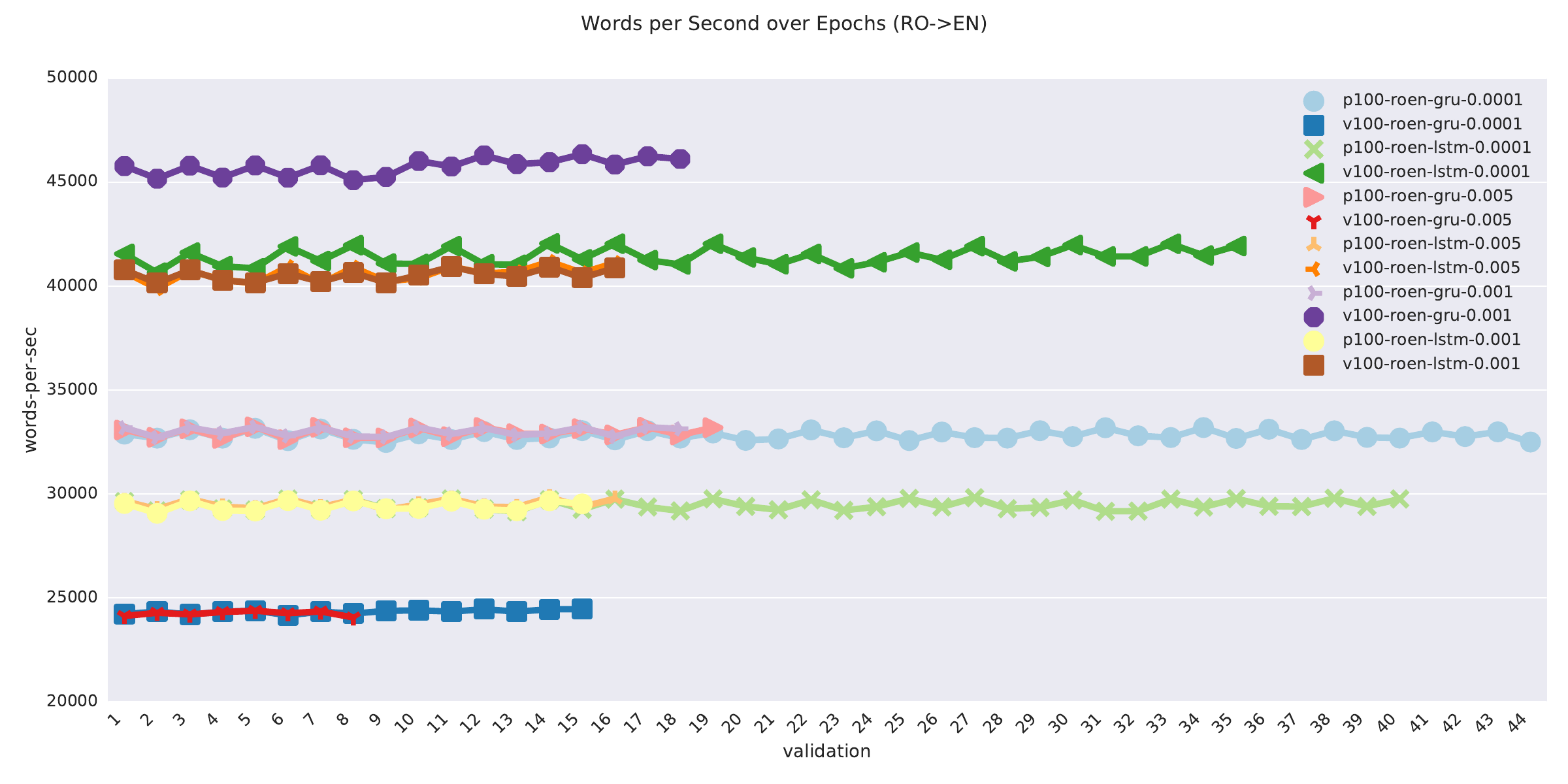}
\caption{Average words-per-second for the RO $\rightarrow$ EN translation task, comparing systems.}
\label{fig:wps}
\end{figure*}

\subsection{Convergence Speed}

%\begin{table} 
%%\tiny
%\caption{Words-per-second over epochs for the RO $\rightarrow$ EN translation task, comparing activation units, learning rates and GPUs.}
%\centering 
%\begin{tabular}{|c|r|cccc|} \hline 
%\ & & \multicolumn{2}{c}{words-per-sec (avg)}  & \multicolumn{2}{c|}{validation} \\
%\ cell & learn-rt & P100 & V100 & P100 & V100 \\ \hline \hline
% \ GRU & 1e-3 & 33009.23 & \cellcolor{gray!25}45762.54 & 18000 &\cellcolor{gray!25}18000  \\ %\hline
%\ & 5e-3 & 32965.23 & 24253.14 & 19000 & 8000 \\ %\hline
%\ & 1e-4 & 32828.61 & 24341.96 & 44000 & 16000 \\ \hline
%\ LSTM & 1e-3 & 29412.87 & 40534.06 & 15000 & 16000  \\
%\ & 5e-3 & 29536.65 & 40598.24 & 16000 & 16000  \\
%\ & 1e-4 & 29478.51 & \cellcolor{gray!25}41441.37 & 40000 & \cellcolor{gray!25}35000 \\ \hline
%\end{tabular}
%\label{tab:wps}
%\end{table}

\begin{table*} 
%\tiny
\caption{Total training time for four translation directions, comparing systems.}
\centering 
\begin{tabular}{|c|r|cccc|cccc|} \hline 
\ & & \multicolumn{2}{c}{ro$\rightarrow$en}  & \multicolumn{2}{c|}{en$\rightarrow$ro} & \multicolumn{2}{c}{de$\rightarrow$en}  & \multicolumn{2}{c|}{en$\rightarrow$de} \\
\ cell & learn-rt & P100 & V100 & P100 & V100 & P100 & V100 & P100 & V100 \\ \hline \hline
 \ GRU & 1e-3 & 8:48 & 6:21 & \cellcolor{gray!25}7:47 & \cellcolor{gray!25}5:26 & 18:47 & 19:40 & \cellcolor{gray!25}9:26 & 6:41 \\ %\hline
\ & 5e-3 & 9:41 & \cellcolor{gray!25}4:52 & 8:38 & 6:02 & 23:57 & \cellcolor{gray!25}4:36 & n/a & 10:56\\ %\hline
\ & 1e-4 & 21:58 & 9:43 & 12:33 & 8:59 & 23:50 & 21:09 & dnf & 23:58\\ \hline
\ LSTM & 1e-3 & \cellcolor{gray!25}8:10 & 6:34 & 7:49 & 5:36 & \cellcolor{gray!25}16:33 & 13:39 & 13:50 & 12:24 \\
\ & 5e-3 & 9:02 & 6:34 & 10:44 & 8:32 & n/a & 5:12 & 9:37 & \cellcolor{gray!25}4:35 \\
\ & 1e-4 & 22:29 & 14:05 & 13:46 & 9:45 & n/a & 23:57 & dnf & dnf \\ \hline
\end{tabular}
\label{tab:time}
\end{table*}

Figure~\ref{fig:wps} shows the average words-per-second for the RO $\rightarrow$ EN translation task, comparing systems.   The average words-per-second executed remained consistent across epochs.  The system that was able to achieve the most words-per-second was v100-roen-gru-0.001, whereas the one that achieved the least words-per-second was the v100-roen-gru-0.005.  Surprisingly, the best and worst performer was the v100-roen-gru, depending on its learning rate, with the sweet spot at 0.001.  This confirms 0.001 as the best learn rate that can execute a decent number of words-per-second and achieve a fairly high accuracy, as evident in previous studies, across all systems.

Table~\ref{tab:wps_ro} also displays words-per-second and validation, comparing activation units, learning rates and GPUs.  When fixing learning rate, the V100 was able to execute more words-per-second than the P100, and was able to converge at an earlier iteration.  When comparing hidden units, GRUs were able to execute higher words per second on a GPU and converge at a reasonable rate (at 18000 iterations) for most learning rates, except for 5e-3.  When looking at LSTMs, words-per-second executed on a V100 was similar, although at a higher learning rate it was able to converge at 42000 iterations, but at the cost of longer training time and slower convergence (35000 iterations).
%
%\begin{table*} 
%%\small
%\caption{Average time spent per iteration for RO $\rightarrow$ EN and EN $\rightarrow$ RO translation directions, comparing systems, with standard deviation in parenthesis and epochs in brackets.}
%\centering 
%\begin{tabular}{|c|r|P{2.5cm}P{2.5cm}|P{2.5cm}P{2.5cm}|} \hline 
%\ & & \multicolumn{2}{c|}{ro$\rightarrow$en}  & \multicolumn{2}{c|}{en$\rightarrow$ro} \\
%\ cell & learn-rt & P100 & V100 & P100 & V100 \\ \hline \hline
% \ GRU & 1e-3 & 1807.362941 (142.43) [17] & 1304.076471 (102.67) [17]
% & 1829.790714 (166.06) [14] & 1278.770714 (117.63) [14] \\ %\hline
%\ & 5e-3 & 1814.640556 (140.01) [18] & 2472.531429 (11.16) [7] & 1816.642500 (165.40) [16] & 2385.243333 (15.08) [9] \\ %\hline
%\ & 1e-4 & 1823.828837 (129.08) [43] & 2466.306429 (11.29) [14] & 1839.624583 (167.28) [24] & 2369.436923 (13.79) [23] \\ \hline
%\ LSTM & 1e-3 & 2032.362857 (155.58) [14] & 1470.278 (108.79) [15] & 2010.199231 (146.74) [13] & 1438.945385 (107.76) [13]  \\
%\ & 5e-3 & 2018.048 (148.21) [15] & 1469.054 (110.05) [15] & 2014.716667 (144.41) [18] & 1474.787500 (100.57) [20] \\
%\ & 1e-4 & 2026.976154 (147.46) [39] & 1445.585882 (106.30) [34] & 2037.517083 (140.28) [24] & 1443.758333 (99.68) [24] \\ \hline
%\end{tabular}
%\label{tab:avg_roen}
%\end{table*}

Table~\ref{tab:time} shows the corresponding total training time for the four translation directions, comparing GPUs, activation units, and learning rates.  The dropout rate was set at 0.2, which was the best performer in most cases (Tab~\ref{tab:dropout}).  Table~\ref{tab:time} shows that the training time increased as the learning rates were decreased.  In general, Romanian took a fraction of the time to complete training (usually under 10 hours), whereas German took 18-22 hours to complete training.

\subsection{Cost of Tuning a Hyper-Parameter}
Table~\ref{tab:avg_roen} displays the average time spent per epoch for the Romanian $\leftrightarrow$ English translation task, and Table~\ref{tab:avg_deen} displays the average time spent per epoch for the German $\leftrightarrow$ English translation task, comparing learning rates, activation cells, and GPUs.  The mean is displayed in each cell, with the standard deviation in parenthesis and the number of epochs executed in brackets.  For both tasks, dropout was set to 0.2.  Surprisingly, GRUs take longer on the V100 on average with larger learning rates (5e-3, 1e-4) vs the P100, whereas for LSTMs, the V100s clearly speeds up execution per epoch.  Note also that the learning rate does not have a significant change in the average time spent per epoch, except for the case with GRUs  executing on the V100 with large learning rates.   The learning rate does have an effect on the number of epochs executed, as seen in brackets as the learning rate increases.  Table~\ref{tab:avg_deen} reports on the German $\leftrightarrow$ English translation tasks.  The same observation can be made for this task, where GRUs spend less time per epoch compared to LSTMs, and that the average time spent per epoch remains fixed as the learnignrate increases.

\section{Summarize Findings}
This work reveals the following, with respect to tuning hyper-parameters:

\begin{itemize}
\item Dropout is neccessary to avoid overfitting.  The recommended probability rate is 0.2 to 0.3.
\item LSTMs take longer than GRUs per epoch, but achieves better accuracy.
\item Although the average time spent per epoch remains fixed as learning rates increase, the total number of epochs executed per training run increases as the learning rates increase.
\item Tensor core GPUs, particularly the V100, provide more words that can be processed per second, compared to non-tensor core GPUs, such as the Pascal P100.
\end{itemize}

\begin{table*} 
%\small
\caption{Average time spent per iteration for RO $\rightarrow$ EN and EN $\rightarrow$ RO translation directions, comparing systems, with standard deviation in parenthesis and epochs in brackets.}
\centering 
\begin{tabular}{|c|r|P{2.5cm}P{2.5cm}|P{2.5cm}P{2.5cm}|} \hline 
\ & & \multicolumn{2}{c|}{ro$\rightarrow$en}  & \multicolumn{2}{c|}{en$\rightarrow$ro} \\
\ cell & learn-rt & P100 & V100 & P100 & V100 \\ \hline \hline
 \ GRU & 1e-3 & 1807.362941 (142.43) [17] & 1304.076471 (102.67) [17]
 & 1829.790714 (166.06) [14] & 1278.770714 (117.63) [14] \\ %\hline
\ & 5e-3 & 1814.640556 (140.01) [18] & 2472.531429 (11.16) [7] & 1816.642500 (165.40) [16] & 2385.243333 (15.08) [9] \\ %\hline
\ & 1e-4 & 1823.828837 (129.08) [43] & 2466.306429 (11.29) [14] & 1839.624583 (167.28) [24] & 2369.436923 (13.79) [23] \\ \hline
\ LSTM & 1e-3 & 2032.362857 (155.58) [14] & 1470.278 (108.79) [15] & 2010.199231 (146.74) [13] & 1438.945385 (107.76) [13]  \\
\ & 5e-3 & 2018.048 (148.21) [15] & 1469.054 (110.05) [15] & 2014.716667 (144.41) [18] & 1474.787500 (100.57) [20] \\
\ & 1e-4 & 2026.976154 (147.46) [39] & 1445.585882 (106.30) [34] & 2037.517083 (140.28) [24] & 1443.758333 (99.68) [24] \\ \hline
\end{tabular}
\label{tab:avg_roen}
\end{table*}

\begin{table*} 
% \small
\caption{Average time spent per iteration for DE $\rightarrow$ EN and EN $\rightarrow$ DE translation directions, comparing systems, with standard deviation in parenthesis and epochs in brackets.}
\centering 
\begin{tabular}{|c|r|P{2.5cm}P{2.5cm}|P{2.5cm}P{2.5cm}|} \hline 
\ & & \multicolumn{2}{c|}{de$\rightarrow$en}  & \multicolumn{2}{c|}{en$\rightarrow$de} \\
\ cell & learn-rt & P100 & V100 & P100 & V100 \\ \hline \hline
 \ GRU & 1e-3 & 3430.330526 (124.58) [19] & 2555.738148
(95.76) [27] & 3432.534444 (128.70) [9] & 2535.11 (88.61) [9] \\ %\hline
\ & 5e-3 & 3450.174167 (133.13) [24] & 4898.036667 (47.79) [3] & n/a & 2432.112000 (87.91) [15] \\ %\hline
\ & 1e-4 & 3425.231600 (129.98) [25] & 4907.070667 (51.24) [15] & n/a & 2434.452000 (90.02)  [35] \\ \hline
\ LSTM & 1e-3 & 3887.889333 (164.183) [15] & 2898.554375 (129.37) [16] & 3840.552500 (162.85) [12] & 2761.088125 (116.41) [16]  \\
\ & 5e-3 & 3855.21 (162.27) [13] & 2852.335 (121.95) [6] & 3859.903750 (167.48) [8] & 2886.194 (122.26) [5] \\
\ & 1e-4 & n/a & 2814.689000 (118.66) [30] & n/a & n/a \\ \hline
\end{tabular}
\label{tab:avg_deen}
\end{table*}

\section{Discussion}
The variation in the results, in terms of language translation, hyper-parameters, words-per-second executed and BLEU scores, in addition to the hardware the training was executed on demonstrates the complexity in learning the grammatical structure between the two languages.  In particular, the learning rate set for training, the hidden unit selected for the activation function, the optimization criterion and the amount of dropout applied to the hidden connections all have a drastic effect on overall accuracy and training time.  Specifically, we found that a lower learning rate achieved the best performance in terms of convergence speed and BLEU score.  Also, we found that the V100 was able to execute more words-per-second than the P100 in all cases.  When looking at accuracy as a whole, LSTM hidden units outperformed GRUs in all cases.  Lastly, the amount of dropout applied on a network in all cases prevented the model from overfitting and achieve a higher accuracy.

The multidimensionality of hyper-parameter optimization poses a challenge in selecting the architecture design for training NN models, as illustrated by the varying degrees of behavior across systems and its performance outcome.  This work investigated how the varying design decisions can affect training outcome and provides neural network designers how to best look at which parameters affect performance, whether accuracy, words processed per second, and convergence expectation.  Coupled with massive datasets for parallel text corpuses and commodity heterogenous GPU architectures, the models trained were able to achieve WMT grade accuracy with the proper selection of hyper-parameter tuning.

\section{Conclusion}
We analyzed the performance of various hyper-parameters for training a NMT, including the optimization strategy, the learning rate, the activation cell, and the GPU across various systems for the WMT 2016 translation task in four translation directions.  Results demonstrate that a proper learning rate and a minimal amount of dropout is able to prevent overfitting as well as achieve high training accuracy.  

Future work includes developing optimization methods to evaluate how to best select hyper-parameters.  By statically analyzing the computational graph that represents a NN in terms of instruction operations executed and resource allocation constraints, one could derive execution performance for a given dataset without running experiments.

\bibliographystyle{aaai}
\bibliography{nmt}

\end{document}